# First Competitive Ant Colony Scheme for the CARP

**Lacomme Philippe**[1]    **Christian Prins**[2]
**Alain Tanguy**[1]

Research Report LIMOS/RR-04-21

Juillet 2004

[1] Université Blaise Pascal
Laboratoire d'Informatique (LIMOS) UMR CNRS 6158,
Campus des Cézeaux, 63177 Aubiere Cedex
lacomme@sp.isima.fr, tanguy@sp.isima.fr

[2] Université de Technologie de Troyes,
ISTIT (équipe OSI) FRE CNRS 2732
12, Rue Marie Curie, BP 2060, F-10010 Troyes Cedex (France)
prins@utt.fr

# Abstract

This paper addresses the Capacitated Arc Routing Problem (CARP) using an Ant Colony Optimization scheme. Ant Colony schemes can compute solutions for medium scale instances of VRP. The proposed Ant Colony is dedicated to large-scale instances of CARP with more than 140 nodes and 190 arcs to service. The Ant Colony scheme is coupled with a local search procedure and provides high quality solutions. The benchmarks we carried out prove possible to obtain solutions as profitable as CARPET ones can be obtained using such scheme when a sufficient number of iterations is devoted to the ants. It competes with the Genetic Algorithm of Lacomme *et al.* regarding solution quality but it is more time consuming on large scale instances. The method has been intensively benchmarked on the well-known instances of Eglese, DeArmon and the last ones of Belenguer and Benavent. This research report is a step forward CARP resolution by Ant Colony proving ant schemes can compete with Taboo search methods and Genetic Algorithms

# Résumé

Cet article concerne la résolution du Capacitated Arc Routing Problem avec un algorithme de type colonies de fourmis. Il a déjà été prouvé que de tels algorithmes permettaient de résoudre des problèmes de VRP de tailles modestes. L'algorithme proposé a pour ambition de résoudre des instances de CARP de très grandes tailles comportant plus de 140 nœuds et 190 arcs. La méthode proposée est hybridée avec une méthode de recherche locale qui accroît de manière significative la convergence vers de bonnes solutions. Nous montrons sur des exemples de la littérature que la méthode proposée permet d'obtenir des solutions comparables à celle de la méthode CARPET au prix d'un nombre d'itérations relativement élevé. La méthode permet même de concurrencer l'algorithme génétique de Lacomme *et al.* mais elle s'avère plus longue en terme de temps de calcul. Toutes les comparaisons ont été réalisées sur les instances de Eglese, DeArmon et sur les instances de Belenguer et Benavent. Ce rapport de recherche tente de prouver que les techniques d'optimisation à base de colonies de fourmis peuvent concurrencer les méthodes « tabou » ainsi que les algorithmes génétiques.

*Mots-clés : colonies de fourmis, tournées sur arcs*

# 1  Introduction

## *1.1  The Capacitated Arc Routing Problem*

The *Capacitated Arc Routing Problem* (*CARP*) is defined in the literature on an undirected network $G = (V,E)$ with a set $V$ of $n$ nodes and a set $E$ of $m$ edges. A fleet of identical vehicles of capacity $Q$ is based at a depot node $s$. A subset $R$ of edges requires service by a vehicle. All edges can be traversed any number of times. Each edge $i$ has a traversal cost $c_i > 0$ and a demand $q_i \geq 0$. The CARP consists on determining a set of vehicle trips with minimum total cost. Each trip starts and ends at the depot, each required edge is serviced by one single trip, and the total demand handled by any vehicle does not exceed $Q$. The cost of a trip is the sum of the costs of its serviced edges and of its intermediate connecting paths.

Golden B-L. and R.T.Wong [1], Benavent E., V. Campos and A. Corberan, E. Mota [2], Belenguer J.M. and E. Benavent [3] have investigated integer linear programming formulations and they have proposed lower bounds for the CARP. Since exact algorithms (like the branch-and-bound method of Hirabayashi *et al.* [4]) are limited to small instances (30 edges), larger instances must be tackled in practice by heuristic approaches. Powerful greedy heuristics include Path-Scanning [1], improved Construct-Strike [5], Augment-Insert [6], Augment-Merge [7] and Ulusoy's tour splitting method [8]. Concerning metaheuristics, Li [9] applied simulated annealing and taboo search to a road gritting problem, and Eglese [10] designed a simulated annealing approach for a multi-depot gritting problem with side constraints. The most efficient metaheuristics published so far are: a sophisticated taboo search method (CARPET) of Hertz *et al.* [11] and a hybrid genetic algorithm proposed by Lacomme *et al.* [12].

## *1.2  Ant Colony*

Ant Colony schemes have been successfully applied to numerous combinatorial optimization problems including graph coloring [13], Quadratic Assignment Problem (QAP) [14], Traveling Salesman Problem (TSP) [15], Vehicle Routing Problem (VRP) [16, 17, 18], Vehicle Routing Problem with Time Window (VRPTW) [19]. A presentation of Ant Colony for routing is described in [20]. An important result in the field of ant algorithms was published in [21]: Gutjahr gave a formal proof that, under certain conditions, a slightly limited version of the Ant System (called Graph-based Ant System) converges to the optimal solution of the given problem with a probability that can be made arbitrarily close to 1.

In the beginning, no collective memory is used and ants use only heuristic information. Pheromone deposition is proportional to the fitness that can be defined for minimization objective as the inverse of the solution quality or solution cost. As stressed by Donati *et al.* [22] the fitness may be defined in different ways to reflect the optimization objectives: a combination of the travel distance, the travel time, the waiting times and so on. This pheromone trail guides ants in their future decision making paths with high pheromone concentration more attractive. Since pheromone is not permanent but evaporates over time, unused paths become less and less attractive while those frequently used attract more ants. During one iteration, the ants construct one solution based on heuristic scheme and on the pheromone trails. The pheromone trails are updated using the solutions fitness. Local search can also be applied to ant solutions increasing the convergence rate. The process is iterated until a lower bound is reached or a maximal number of iterations is carried out.

The remainder part of the paper is organized as follows. First, an Ant Colony framework with an elitist strategy dedicated to the CARP is proposed. Second, an intensive benchmark is provided to evaluate the method performances and behavior.

## 2  Ant Colony Proposal

### 2.1  *Notations*

$f$      number of ants

$f_e$      number of elitist ants

$\mu$      notation used for one ant, its rank and the solution computed by the ant

$I_{\max}$      maximal number of iterations

$n_s$      number of iterations without improvement before pheromone erasing

$w_{ij}$      shortest path length from the required arc $i$ to the required arc $j$

$M_d$      shortest path of maximal length between two required arcs: $M_d = \underset{i,j}{Max}\, w_{ij}$

$p_{LS}$  probability for one ant to experiment a local search procedure

$p_p$  probability to ignore pheromone trails for combining required arcs

$k$  maximal size of $\Psi_i^\mu$ and $\Omega_i^\mu$

$\alpha, \beta$  relative influence of criteria (saving measure of moving to another required arc and pheromone attraction)

$F^\mu$  the weight affected to the ant number $\mu$

$\rho$  the trail persistence, $0 \leq \rho \leq 1$

$s_{ij}$  saving measure of moving from the required arc $i$ to the required arc $j$: $s_{ij} = (M_d - w_{ij})/M_d$

$\tau_{ij}$  existing amount of pheromone from the required arc $i$ to the required arc $j$

$\Delta\tau_{ij}^\mu$  deposit amount of pheromone laid by ant number $\mu$ when moving from required arc $i$ to required arc $j$

$L^\mu$  current value of the solution found by the ant number $\mu$

$t_\mu$  taboo list of ant $\mu$

$\Omega_i^\mu$  set of required arcs not in $t_\mu$ and yielding the best savings

$\Psi_i^\mu$  set of required arcs not in $t_\mu$ and yielding the best pheromone level

$P_{ij}^\mu$  probability, for the ant $\mu$, to combine the required arcs $i$ and $j$

## 2.2 Ant Colony Framework

The network is stored as a directed internal graph using two opposite arcs per edge and one dummy loop on the depot. The nodes are dropped out and an arc index list is used. In any solution, each trip is stored as a sequence of required arcs with a total cost and a total load. Shortest paths between tasks are not stored. The cost of a trip is the collecting costs of its required arcs plus the traversal costs of its intermediate paths. The graph and solutions are represented according to the data structure described in [12] and with respect to the proposal of Lacomme *et al.*, the solutions are giant tours with no trip delimiters sorted in increasing cost order. The giant tours are split into solutions regarding the vehicle constraints, using the Split procedure.

The ant system framework consists in the following steps (figure 1):

1. generation of solutions by powerful constructive heuristics dedicated to CARP;
2. generation of solutions by ants according to pheromone information;
3. application of a local search to the ant solutions with a fixed probability;
4. updating the pheromone information;
5. iterate steps 2 to 4 until the lower bound or some completion criteria are reached.

```
Generation of the initial set of f solutions
I_c = 1
Repeat
    Pheromone trails deposit
    For µ := 1 to f do
        Repeat
            Select a required arc i to be serviced next
            Add i in the current solution under construction
            Update the taboo list t_µ of the ant µ
        until ant µ has completed a tour
        with probability P_LS, apply Local Search
        Calculate the solution cost
        If µ is not elitist Then
            save the solution whatever the cost
        Else
            save solution only if a better tour is obtained by ant µ
        EndIf
    EndDo
    Sort the ants in decreasing cost order
        If for ant f no improvement occurs during n_s iterations Then
            Erase the pheromone trails
        EndIf
    I_c = I_c + 1
Until (the lower bound is reached) or ( I_c = I_max )
```

***Figure 1.*** *Ant Colony algorithm*

The population is divided in two sets: $f_e$ elitist ants and $f - f_e$ non-elitist ants. All ants start at the anthill: depot node. Note that for VRP, several authors promote initial assignment of ants at each customer node at the beginning of iterations which implies that the number of ants is equal to the number of customers [22]. The elitist ants tend to favor the convergence of the algorithm and the non-elitist ones attempt to control the diversification process. Whatever the solution cost found by a non-elitist ant, it is stored and replaces the previous one. For the elitist ants, the solution is replaced only if it is more promising. To decrease the probability of being captive in a local minimum, the pheromone is erased when $n_s$ iterations have been performed without improvement (figure 1).

## 2.3 Generation of Initial Solutions

Three well-known CARP heuristics have been used: Path-Scanning [7], Augment-Merge [1] and Ulusoy's tour splitting method [8]. The initial set of solutions is completed by random feasible solutions. Each solution is composed of one giant tour without the vehicle capacity constraint. This giant tour is a list of required arcs linked by shortest paths and the Split procedure [12] breaks this tour optimally into trips. Path Scanning algorithm builds one trip at a time. In constructing each trip, the sequence of arcs is extended by joining the arcs looking the most promising ones, until the vehicle capacity is reached. Possible criteria are minimization of the distance, maximization of productivity. Augment-Merge is composed of two phases. First, each required arc is serviced by a separate trip. Second, Augment considers the trips one by one, starting with the longest one and evaluates the concatenation of trips yielding the largest saving. Ulusoy's algorithm is composed of two steps. The first step relaxes capacity to build a giant tour $S$ containing the required arcs. In a second step, $S$ is optimally split into trips under the vehicle capacity constraint using Ulusoy's algorithm.

## 2.4 Solution Improvement

The pheromone update is done using the well-known formula:

$$\tau_{ij} \leftarrow \rho.\tau_{ij} + \sum_{\mu=1}^{n} \Delta\tau_{ij}^{\mu} \text{ with } \Delta\tau_{ij}^{\mu} = F^{\mu}/L^{\mu} \qquad (1)$$

The contribution level of this global information depends on the quality of the solution. The basic weight $F^{\mu} = 1$ denotes that no ant is considered more promising than another one and the same weight (value 1) is affected to each ant. To have a quantity of pheromone laid by ants depending on their rank $\mu$ it is possible to choose: $F^{\mu} = \mu$. The solution we promote is to consider some graph properties and specially the maximal distance between two required arcs. One can suppose the objective function is more "chaotic" when large distances occur in the network and it is possible to link $F^{\mu}$ and $M_d$ according to the following formula:

$$F^{\mu} = \mu \times (M_d - 1)/(f - 1) + (f - M_d)/(f - 1) \quad (2)$$

With probability $p_p$, its next required arc is chosen taking into account the unproductive shortest path from its current position to another required arc. The following formulae are applied respectively with probability $p_p$ and $1 - p_p$:

$$P_{ij}^{\mu} = \begin{cases} 1/k & if\ j \in \Omega_i^{\mu} \\ 0 & otherwise \end{cases} \quad (3) \qquad P_{ij}^{\mu} = \begin{cases} \dfrac{[s_{ij}]^{\alpha}[\tau_{ij}]^{\beta}}{\sum_{k \in \Psi_i^{\mu}}[s_{ij}]^{\alpha}[\tau_{ij}]^{\beta}} & if\ j \in \Psi_i^{\mu} \\ 0 & otherwise \end{cases} \quad (4)$$

The first one is similar to the ranking strategy described, for instance, by Bullnheimer [23]. The second one uses pheromone trails.

### 2.5 Local Search

To improve the performances of metaheuristics it is a common practice to include a local search procedure. Previous published works on Ant Colony for the VRP prove that local search coupled with the Ant Colony method considerably improves the solutions quality (see for example [22]). The local search scheme dedicated to the CARP and proposed by Lacomme *et al.* [12] is used to improve solution with probability $p_{LS}$. It is an iterative improvement procedure based on three moves:

1. remove one required arc and reinsert it at another location;
2. remove two consecutive required arcs and reinsert them at another location;
3. two-opt moves.

The local search algorithm detects and performs the first feasible and improving move. This process is iterated until no such move is found. The split procedure [8] is applied to get the solution cost.

# 3 Numerical Evaluation

In this section, we present numerical experiments for the proposed Ant Colony scheme compared to the best methods for the CARP including CARPET and the Genetic Algorithm [12, 24]. The experiments were carried out on a Pentium III 800 MHz under Windows 2000. The scheme has been implemented using Delphi 6.

## 3.1 Instances and Ant Colony Scheme Parameters

The instance and parameters setting is summarized up in table 1. The following extra notations are introduced: $\tau$ the number of required arcs and $n$ the number of nodes.

**Table 1.** *Instances and parameters*

| | | |
|---|---|---|
| $f = 60$ | $k = 10$ | $I_{max} = 200$ |
| $f_e = 10$ | $P_{LS} = 50\%$ | $n_s = 10$ |
| $\rho = 0.90$ | $p_p = 10\%$ | $\alpha = \beta = 1$ |

The benchmark has been performed using the well-known instances of DeArmon, Eglese, Belenguer and Benavent using the parameters values presented in table 1. For each instance, the notations of table 2 are used. To evaluate the performances three experiments have been carried out with different seeds for the random generator. Results over three restarts are in column BACO (Best Ant Colony Optimization).

**Table 2.** *Notations used in table of results*

| | |
|---|---|
| LB | Lower Bound |
| C | CARPET algorithm of Hertz |
| GW | Golden and Wang's algorithm |
| AM | Augment Merge heuristic |
| UL | Ulusoy |
| Ant | Ant Colony algorithm |
| Time | computation time for I iterations |
| BACO | best results over three experiments |
| Avg Time | average computational time for BACO |
| Dev | deviation regarding the LB, for a cost $x$: $Dev = \dfrac{x - LB}{LB}$ |
| Av.Dev | average deviation regarding the LB |
| Nb hits | number of optima proved by the method |
| I | iteration at which the best GA value is found |
| GA | Genetic Algorithm |

Grey boxes show solutions equal or better than CARPET and bold denote solutions equal of better than Genetic Algorithm ones. Note Lacomme *et al.* propose a GA with restarts denoted *Std MA* in [24] and results under various settings (*Best MA*). The GA column in tables refers to *Std GA* of [24].

## 3.2 Numerical Experiments

### 3.2.1 Results on DeArmon's Instances

The Ant Colony Optimization scheme (table 3) outperforms CARPET for the three experiments. Whatever the experiments, the deviation is about 0.34% that is better than CARPET deviation (0.50%). Let us note that whatever the experiment, for the Gdb8 instance the lower bound is not reached but the value 350 is better than CARPET solution (value 352) and equal to the solution of the Genetic Algorithm except for Gdb9, Gdb13 and Gdb23.

*Ant System configuration over iterations.*

The initial population is composed of random solutions and three heuristic solutions. Because no optimization has been applied on the initial population, the ant population is well spread from 360 (best heuristic solution) to 800 (worst random solution). The generation of the initial set of solutions provides a high diversification of solutions. Figure 2 gives a graphical representation of ants sorted by increasing cost after 20 iterations. The convergence of the algorithm produces stepwise modifications of the cost distribution of the ant population and the convergence is initialized.

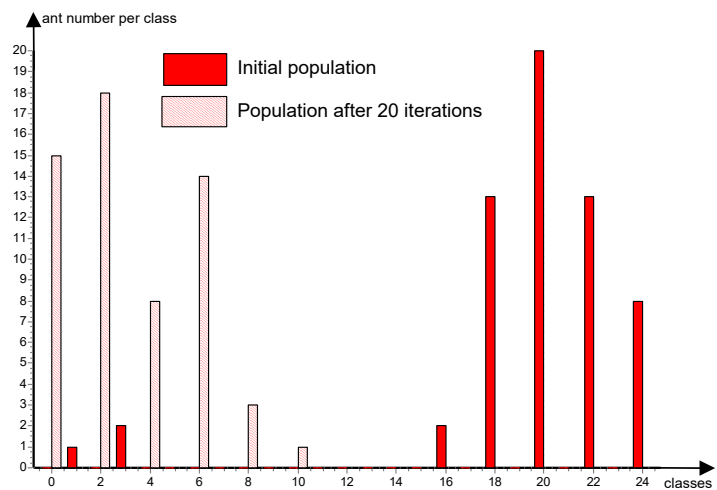

***Figure 2.*** *Population distribution after 20 iterations (Gdb9)*

*Table 3*. DeArmon's instances

|  |  |  |  |  |  |  |  |  |  |  | Exp.1 |  |  |  | Exp.2 |  |  |  | Exp.3 |  |  |  |  |
|---|---|---|---|---|---|---|---|---|---|---|---|---|---|---|---|---|---|---|---|---|---|---|---|---|
| FILE | n | τ | LB | C | Dev | GA | Dev | GW | AM | UL | Ant | Dev | Time | I | Ant | Dev | Time | I | Ant | Dev | Time | I | BACO | Dev |
| gdb1 | 12 | 22 | 316 | 316 | 0.00 | 316 | 0.00 | 350 | 349 | 330 | **316** | 0.00 | 0.49 | 1 | **316** | 0.00 | 0.44 | 1 | **316** | 0.00 | 0.44 | 1 | **316** | 0.00 |
| gdb2 | 12 | 26 | 339 | 339 | 0.00 | 339 | 0.00 | 366 | 370 | 353 | **339** | 0.00 | 1.10 | 2 | **339** | 0.00 | 2.15 | 4 | **339** | 0.00 | 2.08 | 4 | **339** | 0.00 |
| gdb3 | 12 | 22 | 275 | 275 | 0.00 | 275 | 0.00 | 293 | 319 | 297 | **275** | 0.00 | 0.49 | 1 | **275** | 0.00 | 0.55 | 1 | **275** | 0.00 | 0.49 | 1 | **275** | 0.00 |
| gdb4 | 11 | 19 | 287 | 287 | 0.00 | 287 | 0.00 | 287 | 302 | 320 | **287** | 0.00 | 0.05 | 0 | **287** | 0.00 | 0.05 | 0 | **287** | 0.00 | 0.06 | 0 | **287** | 0.00 |
| gdb5 | 13 | 26 | 377 | 377 | 0.00 | 377 | 0.00 | 438 | 423 | 407 | **377** | 0.00 | 1.65 | 3 | **377** | 0.00 | 4.18 | 5 | **377** | 0.00 | 0.61 | 1 | **377** | 0.00 |
| gdb6 | 12 | 22 | 298 | 298 | 0.00 | 298 | 0.00 | 324 | 340 | 318 | **298** | 0.00 | 1.48 | 3 | **298** | 0.00 | 1.43 | 2 | **298** | 0.00 | 0.49 | 1 | **298** | 0.00 |
| gdb7 | 12 | 22 | 325 | 325 | 0.00 | 325 | 0.00 | 363 | 325 | 330 | **325** | 0.00 | 0.11 | 0 | **325** | 0.00 | 0.11 | 0 | **325** | 0.00 | 0.06 | 0 | **325** | 0.00 |
| gdb8 | 27 | 46 | 344 | 352 | 2.33 | 350 | 0.02 | 463 | 393 | 388 | **350** | 0.02 | 171.92 | 112 | **350** | 0.02 | 68.87 | 43 | **350** | 0.02 | 151.04 | 100 | **350** | 0.02 |
| gdb9 | 27 | 51 | 303 | 317 | 4.62 | 303 | 0.00 | 354 | 352 | 358 | 306 | 0.01 | 330.1 | 184 | 309 | 0.02 | 125.56 | 66 | 309 | 0.02 | 146.27 | 91 | 306 | 0.01 |
| gdb10 | 12 | 25 | 275 | 275 | 0.00 | 275 | 0.00 | 295 | 300 | 283 | **275** | 0.00 | 0.55 | 1 | **275** | 0.00 | 0.55 | 1 | **275** | 0.00 | 0.94 | 2 | **275** | 0.00 |
| gdb11 | 22 | 45 | 395 | 395 | 0.00 | 395 | 0.00 | 447 | 449 | 413 | **395** | 0.00 | 5.77 | 4 | **395** | 0.00 | 9.06 | 6 | **395** | 0.00 | 7.03 | 5 | **395** | 0.00 |
| gdb12 | 13 | 23 | 448 | 458 | 2.23 | 458 | 0.02 | 581 | 569 | 537 | **458** | 0.02 | 1.27 | 3 | **458** | 0.02 | 3.62 | 9 | **458** | 0.02 | 3.35 | 9 | **458** | 0.02 |
| gdb13 | 10 | 28 | 536 | 544 | 1.49 | 536 | 0.00 | 563 | 560 | 552 | 542 | 0.01 | 26.58 | 51 | 544 | 0.01 | 1.15 | 2 | 544 | 0.01 | 0.60 | 1 | 542 | 0.01 |
| gdb14 | 7 | 21 | 100 | 100 | 0.00 | 100 | 0.00 | 114 | 102 | 104 | **100** | 0.00 | 0.44 | 1 | **100** | 0.00 | 0.44 | 1 | **100** | 0.00 | 0.44 | 1 | **100** | 0.00 |
| gdb15 | 7 | 21 | 58 | 58 | 0.00 | 58 | 0.00 | 60 | 60 | 58 | **58** | 0.00 | 0.11 | 0 | **58** | 0.00 | 0.16 | 0 | **58** | 0.00 | 0.17 | 0 | **58** | 0.00 |
| gdb16 | 8 | 28 | 127 | 127 | 0.00 | 127 | 0.00 | 135 | 129 | 132 | **127** | 0.00 | 2.20 | 4 | **127** | 0.00 | 2.04 | 3 | **127** | 0.00 | 15.27 | 29 | **127** | 0.00 |
| gdb17 | 8 | 28 | 91 | 91 | 0.00 | 91 | 0.00 | 93 | 91 | 93 | **91** | 0.00 | 0.17 | 0 | **91** | 0.00 | 0.22 | 0 | **91** | 0.00 | 0.16 | 0 | **91** | 0.00 |
| gdb18 | 9 | 36 | 164 | 164 | 0.00 | 164 | 0.00 | 177 | 174 | 172 | **164** | 0.00 | 1.10 | 1 | **164** | 0.00 | 1.10 | 1 | **164** | 0.00 | 0.99 | 1 | **164** | 0.00 |
| gdb19 | 8 | 11 | 55 | 55 | 0.00 | 55 | 0.00 | 57 | 63 | 63 | **55** | 0.00 | 0.22 | 1 | **55** | 0.00 | 0.28 | 1 | **55** | 0.00 | 0.22 | 1 | **55** | 0.00 |
| gdb20 | 11 | 22 | 121 | 121 | 0.00 | 121 | 0.00 | 132 | 129 | 125 | **121** | 0.00 | 8.30 | 23 | **121** | 0.00 | 4.17 | 11 | **121** | 0.00 | 54.43 | 145 | **121** | 0.00 |
| gdb21 | 11 | 33 | 156 | 156 | 0.00 | 156 | 0.00 | 176 | 163 | 162 | **156** | 0.00 | 3.73 | 5 | **156** | 0.00 | 8.57 | 13 | **156** | 0.00 | 11.81 | 16 | **156** | 0.00 |
| gdb22 | 11 | 44 | 200 | 200 | 0.00 | 200 | 0.00 | 208 | 204 | 207 | 201 | 0.01 | 2.75 | 2 | **200** | 0.00 | 34.22 | 31 | **200** | 0.00 | 4.89 | 4 | **200** | 0.00 |
| gdb23 | 11 | 55 | 233 | 235 | 0.86 | 233 | 0.00 | 251 | 237 | 239 | 235 | 0.01 | 6.15 | 3 | 235 | 0.01 | 2.31 | 1 | 235 | 0.01 | 12.46 | 7 | 235 | 0.01 |
|  |  |  | Av.Dev(%): | 0.50 |  | 0.17 |  |  |  |  | 0.34 |  |  |  | 0.36 |  |  |  | 0.36 |  |  |  |  | 0.30 |
|  |  |  | Nb hits: | 18 |  | 21 |  |  |  |  | 17 |  |  |  | 18 |  |  |  | 18 |  |  |  |  | 18 |

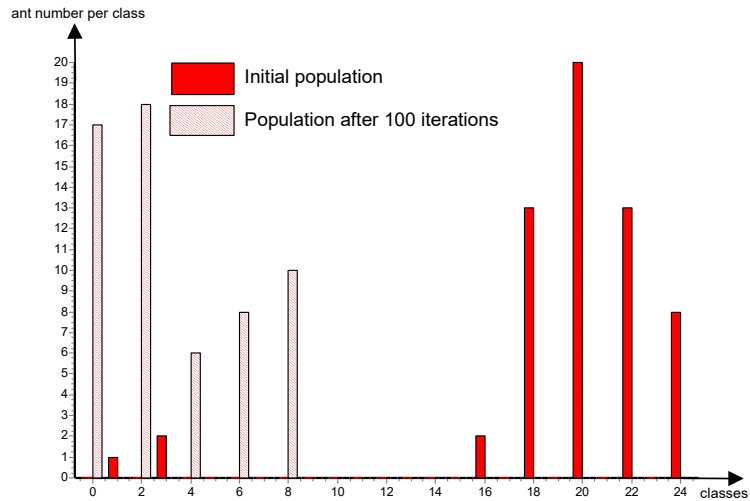

***Figure 3.*** *Population distribution after 100 iterations (Gdb9)*

After 100 iterations, the distribution has changed (figure 3) and ants are concentrated around promising values.

Figure 4 gives the population evolution over iterations. To obtain a legible graphical representation, solutions are sorted in increasing cost order. After 50 iterations, there are 9 solutions better than Carpet one and a global trend appears to minimize the solution cost. Thanks to the non-elitist ants the minimization process continues over iterations (figure 4.c and 4.d) with an important diversity of the cost. After 184 iterations, 10 high quality solutions are identified and the population cost is spread over 320 and 450.

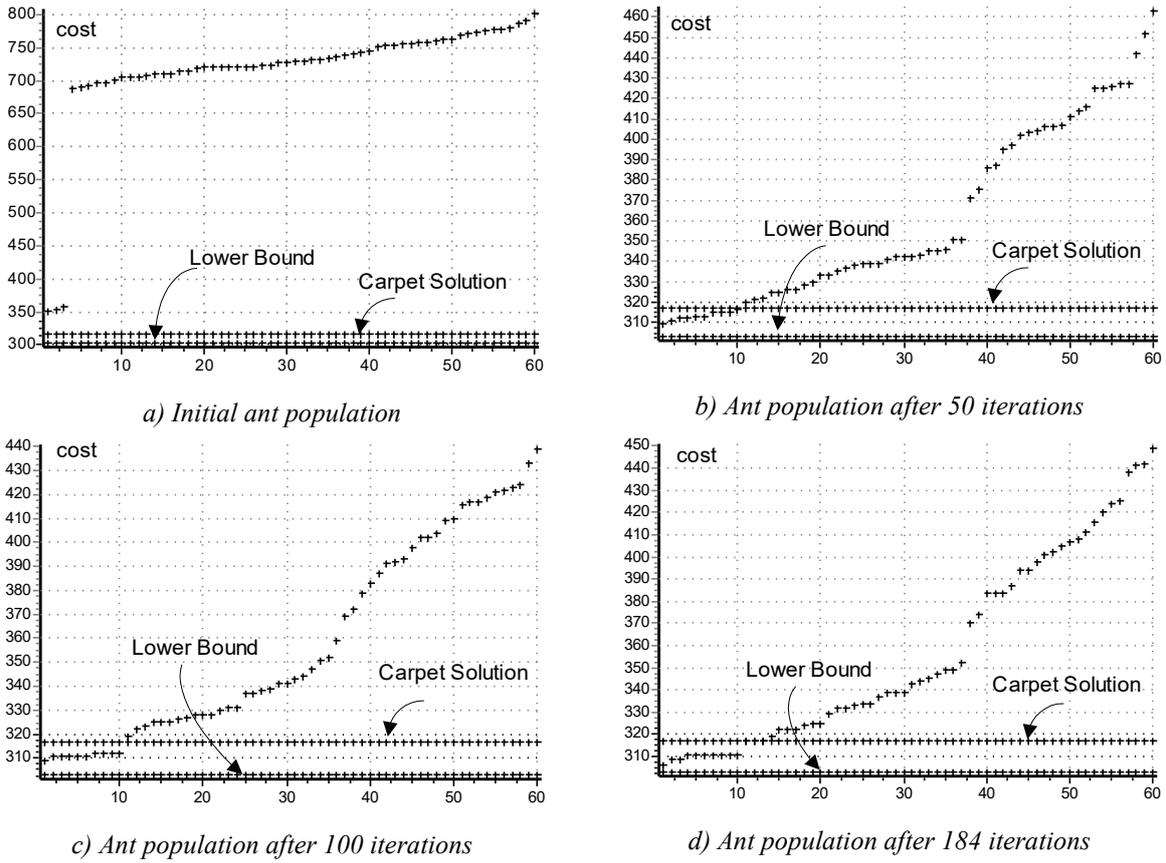

*Figure 4.* Ant population evolution for Gdb9 instance during the first experiment

*Conclusion on DeArmon's instances*

The results prove the efficiency of the ants regarding both the taboo method CARPET and the Genetic Algorithm. For DeArmon's instances (table 3), the ant optimization scheme ever provides solutions equal or better than CARPET solutions and 20 times it provides solutions equal or better than the Genetic Algorithm (table 4).

*Table 4.* DeArmon's instances

| BACO | CARPET | GA |
|---|---|---|
| better than | 3 | 0 |
| equal to | 20 | 20 |
| worse than | 0 | 3 |

### 3.2.2 Results on Belenguer and Benavent's Instances

BACO competes with CARPET (table 5) for almost all instances except for Val8C and Val10b, and provides better deviations than CARPET as regards the lower bound.

*Table 5. Belenguer and Benavent's instances*

| FILE | n | τ | LB | C | Dev | GA | Dev | BACO | Dev | Av. Time |
|---|---|---|---|---|---|---|---|---|---|---|
| val1a | 24 | 39 | 173 | 173 | 0.00 | 173 | 0.00 | **173** | 0.00 | 0.11 |
| val1b | 24 | 39 | 173 | 173 | 0.00 | 173 | 0.00 | **173** | 0.00 | 120.59 |
| val1c | 24 | 39 | 235 | 245 | 0.04 | 245 | 0.04 | **245** | 0.04 | 13.13 |
| val2a | 24 | 34 | 227 | 227 | 0.00 | 227 | 0.00 | **227** | 0.00 | 1.97 |
| val2b | 24 | 34 | 259 | 260 | 0.00 | 259 | 0.00 | **259** | 0.00 | 8.43 |
| val2c | 24 | 34 | 455 | 494 | 0.09 | 457 | 0.00 | **457** | 0.00 | 135.11 |
| val3a | 24 | 35 | 81 | 81 | 0.00 | 81 | 0.00 | **81** | 0.00 | 1.15 |
| val3b | 24 | 35 | 87 | 87 | 0.00 | 87 | 0.00 | **87** | 0.00 | 3.63 |
| val3c | 24 | 35 | 137 | 138 | 0.01 | 138 | 0.01 | **138** | 0.01 | 10.62 |
| val4a | 41 | 69 | 400 | 400 | 0.00 | 400 | 0.00 | **400** | 0.00 | 15.34 |
| val4b | 41 | 69 | 412 | 416 | 0.01 | 412 | 0.00 | **412** | 0.00 | 117.13 |
| val4c | 41 | 69 | 428 | 453 | 0.06 | 428 | 0.00 | 430 | 0.00 | 285.38 |
| val4d | 41 | 69 | 520 | 556 | 0.07 | 541 | 0.04 | **539** | 0.04 | 315.86 |
| Val5a | 34 | 65 | 423 | 423 | 0.00 | 423 | 0.00 | **423** | 0.00 | 49.53 |
| Val5b | 34 | 65 | 446 | 448 | 0.00 | 446 | 0.00 | **446** | 0.00 | 24.30 |
| Val5c | 34 | 65 | 469 | 476 | 0.01 | 474 | 0.01 | **474** | 0.01 | 200.31 |
| val5d | 34 | 65 | 571 | 607 | 0.06 | 581 | 0.02 | 597 | 0.05 | 193.82 |
| val6a | 31 | 50 | 223 | 223 | 0.00 | 223 | 0.00 | **223** | 0.00 | 3.77 |
| val6b | 31 | 50 | 231 | 241 | 0.04 | 233 | 0.01 | **233** | 0.01 | 78.39 |
| val6c | 31 | 50 | 311 | 329 | 0.06 | 317 | 0.02 | **317** | 0.02 | 91.57 |
| val7a | 40 | 66 | 279 | 279 | 0.00 | 279 | 0.00 | **279** | 0.00 | 11.17 |
| val7b | 40 | 66 | 283 | 283 | 0.00 | 283 | 0.00 | **283** | 0.00 | 6.59 |
| val7c | 40 | 66 | 333 | 343 | 0.03 | 334 | 0.00 | **334** | 0.00 | 569.27 |
| val8a | 30 | 63 | 386 | 386 | 0.00 | 386 | 0.00 | **386** | 0.00 | 15.38 |
| val8b | 30 | 63 | 395 | 401 | 0.02 | 395 | 0.00 | **395** | 0.00 | 259.49 |
| val8c | 30 | 63 | 517 | 533 | 0.03 | 527 | 0.02 | 534 | 0.03 | 358.06 |
| val9a | 50 | 92 | 323 | 323 | 0.00 | 323 | 0.00 | **323** | 0.00 | 969.03 |
| val9b | 50 | 92 | 326 | 329 | 0.01 | 326 | 0.00 | **326** | 0.00 | 1076.21 |
| val9c | 50 | 92 | 332 | 332 | 0.00 | 332 | 0.00 | **332** | 0.00 | 1368.47 |
| val9d | 50 | 92 | 382 | 409 | 0.07 | 391 | 0.02 | 404 | 0.06 | 633.98 |
| val10a | 50 | 97 | 428 | 428 | 0.00 | 428 | 0.00 | **428** | 0.00 | 341.81 |
| val10b | 50 | 97 | 436 | 436 | 0.00 | 436 | 0.00 | 437 | 0.00 | 683.42 |
| val10c | 50 | 97 | 446 | 451 | 0.01 | 446 | 0.00 | 448 | 0.00 | 515.79 |
| val10d | 50 | 97 | 524 | 544 | 0.04 | 530 | 0.01 | 538 | 0.03 | 916.10 |
| | | | Av.Dev(%): | | 1.96 | | 0.61 | | 0.90 | |
| | | | Nb hits: | | 15 | | 22 | | 20 | |

CARPET has an average deviation of 1.90%. Over three experiments, the deviation of the ants is 1.11%, 1.05% and 1.04% and BACO has an average deviation of only 0.90%. For the 34 instances, BACO provides solutions equal or better than CARPET for 32 instances (table 6).

*Table 6. Belenguer and Benavent's instances*

| BACO | CARPET | GA |
|---|---|---|
| better than | 17 | 0 |
| equal to | 15 | 27 |
| worse than | 2 | 7 |

The three experiments show that the Ant Colony scheme provides very low deviation as regards lower bound but its deviation is slightly higher than deviation of the Genetic Algorithm. However, BACO competes with the Genetic Algorithm for 27 instances and it is worse for only 7 instances. The average deviation of BACO (0.90%) is close to the Genetic Algorithm deviation (0.61%). The computational time remains acceptable: 20 minutes are required for the last instances of the benchmark (Val9a - Val10d). This computation duration is two times larger than the execution time of the first instances. The first instances are solved in only 2 minutes of computation.

### 3.2.3 Results on Eglese's Instances

Eglese's instances (table 7) are more complex than the previous ones because non-required edges are spread in large-scale instances. For example, instances s1-A to s4-C have 140 required arcs and 190 nodes.

*Table 7.* Eglese's instances

| FILE | n | τ | LB | C | Dev | GA | DEV | GW | AM | UL | BACO | Dev | Av Time |
|---|---|---|---|---|---|---|---|---|---|---|---|---|---|
| e1-A | 77 | 98 | 3515 | 3625 | 0.03 | 3548 | 0.01 | 4115 | 4605 | 3952 | **3548** | 0.01 | 70.68 |
| e1-B | 77 | 98 | 4436 | 4532 | 0.02 | 4498 | 0.01 | 5228 | 5494 | 5054 | 4534 | 0.02 | 307.49 |
| e1-C | 77 | 98 | 5453 | 5663 | 0.04 | 5595 | 0.03 | 7240 | 6799 | 6166 | 5647 | 0.04 | 159.12 |
| e2-A | 77 | 98 | 4994 | 5233 | 0.05 | 5018 | 0.00 | 6458 | 6253 | 5716 | **5018** | 0.00 | 470.39 |
| e2-B | 77 | 98 | 6249 | 6422 | 0.03 | 6340 | 0.01 | 7964 | 7923 | 7080 | 6401 | 0.02 | 406.39 |
| e2-C | 77 | 98 | 8114 | 8603 | 0.06 | 8415 | 0.04 | 10313 | 10453 | 9338 | 8498 | 0.05 | 707.39 |
| e3-A | 77 | 98 | 5869 | 5907 | 0.01 | 5898 | 0.00 | 7454 | 7350 | 6723 | 5934 | 0.01 | 609.83 |
| e3-B | 77 | 98 | 7646 | 7921 | 0.04 | 7822 | 0.02 | 9900 | 9244 | 8713 | 7915 | 0.04 | 781.88 |
| e3-C | 77 | 98 | 10019 | 10805 | 0.08 | 10433 | 0.04 | 12672 | 12556 | 11641 | **10402** | 0.04 | 226.66 |
| e4-A | 77 | 98 | 6372 | 6489 | 0.02 | 6461 | 0.01 | 7527 | 7798 | 7231 | 6520 | 0.02 | 616.78 |
| e4-B | 77 | 98 | 8809 | 9216 | 0.05 | 9021 | 0.02 | 10946 | 10543 | 10223 | 9234 | 0.05 | 839.79 |
| e4-C | 77 | 98 | 11276 | 11824 | 0.05 | 11779 | 0.04 | 13828 | 13623 | 13165 | 11883 | 0.05 | 799.26 |
| s1-A | 140 | 190 | 4992 | 5149 | 0.03 | 5018 | 0.01 | 6382 | 6143 | 5636 | 5049 | 0.01 | 1010.53 |
| s1-B | 140 | 190 | 6201 | 6641 | 0.07 | 6435 | 0.04 | 8631 | 7992 | 7086 | 6541 | 0.05 | 2899.76 |
| s1-C | 140 | 190 | 8310 | 8687 | 0.05 | 8518 | 0.03 | 10259 | 10338 | 9572 | 8561 | 0.03 | 2388.90 |
| s2-A | 140 | 190 | 9780 | 10373 | 0.06 | 9995 | 0.02 | 12344 | 11672 | 11475 | 10368 | 0.06 | 4108.04 |
| s2-B | 140 | 190 | 12886 | 13495 | 0.05 | 13174 | 0.02 | 16386 | 15178 | 14845 | 13676 | 0.06 | 5377.59 |
| s2-C | 140 | 190 | 16221 | 17121 | 0.06 | 16795 | 0.04 | 20520 | 19673 | 19290 | 17115 | 0.06 | 3099.32 |
| s3-A | 140 | 190 | 10025 | 10541 | 0.05 | 10296 | 0.03 | 13041 | 11957 | 11956 | 10619 | 0.06 | 1392.07 |
| s3-B | 140 | 190 | 13554 | 14291 | 0.05 | 14053 | 0.04 | 17377 | 15891 | 15663 | 14264 | 0.05 | 6568.64 |
| s3-C | 140 | 190 | 16969 | 17789 | 0.05 | 17297 | 0.02 | 21071 | 19971 | 20064 | 17797 | 0.05 | 3160.04 |
| s4-A | 140 | 190 | 12027 | 13036 | 0.08 | 12442 | 0.03 | 15321 | 14741 | 13978 | 12868 | 0.07 | 8919.24 |
| s4-B | 140 | 190 | 15933 | 16924 | 0.06 | 16531 | 0.04 | 19860 | 19172 | 18612 | 17090 | 0.07 | 6360.03 |
| s4-C | 140 | 190 | 20179 | 21486 | 0.06 | 20832 | 0.03 | 25921 | 24175 | 23727 | 21314 | 0.06 | 4911.44 |
| | | Av. Dev(%): | | | 4.74 | | 2.47 | | | | | 4.11 | |
| | | Nb hits: | | | 0 | | 0 | | | | | 0 | |

The Augment-Merge method, Golden and Wang's heuristic and Ulusoy's algorithm performances strongly decrease for these instances (see [12, 24] for details). On DeArmon's instances, the lower bound is reached by these heuristics for some instances. This remark remains true for Belenguer and Benavent's instances (Val1a). For Eglese's instances, the deviation of the heuristics as regards the lower bound is high. CARPET provides an average deviation of 4.74%. BACO provides 15 solutions equal or better than CARPET ones and 9 solutions worse than CARPET ones (table 8). The Genetic Algorithm [12, 24] provides high quality solutions and the average deviation is low: 2.47%. The Ant Colony scheme provides an average deviation of 4.11% for BACO, which is two times worse than the Genetic Algorithm deviation.

*Table 8.* Eglese's instances

| BACO | CARPET | GA |
|---|---|---|
| **better than** | 15 | 0 |
| **equal to** | 0 | 3 |
| **worse than** | 9 | 21 |

Significant enlargement of iterations improves solutions quality. For the small scale Eglese's instances, with 1000 iterations the Ant Colony scheme provides an average deviation of 2.94% which is close to the 2.15% of the Genetic Algorithm and better than the 3.84% of the CARPET algorithm (table 9).

*Table 9. Resolution of the small scale Eglese's instances with 1000 iterations (one experiment)*

| FILE | n | τ | LB | C | Dev | GA | DEV | Ant | Dev | Time | I |
|---|---|---|---|---|---|---|---|---|---|---|---|
| e1-A | 77 | 98 | 3515 | 3625 | 0.03 | 3548 | 0.01 | **3548** | 0.01 | 50.59 | 17 |
| e1-B | 77 | 98 | 4436 | 4532 | 0.02 | 4498 | 0.01 | 4514 | 0.02 | 622.86 | 214 |
| e1-C | 77 | 98 | 5453 | 5663 | 0.04 | 5595 | 0.03 | 5632 | 0.04 | 2204.31 | 757 |
| e2-A | 77 | 98 | 4994 | 5233 | 0.05 | 5018 | 0.00 | **5018** | 0.00 | 180.42 | 50 |
| e2-B | 77 | 98 | 6249 | 6422 | 0.03 | 6340 | 0.01 | 6406 | 0.02 | 2641.44 | 742 |
| e2-C | 77 | 98 | 8114 | 8603 | 0.06 | 8415 | 0.04 | 8479 | 0.05 | 3023.08 | 852 |
| e3-A | 77 | 98 | 5869 | 5907 | 0.01 | 5898 | 0.00 | 5902 | 0.01 | 4173.31 | 965 |
| e3-B | 77 | 98 | 7646 | 7921 | 0.04 | 7822 | 0.02 | 7853 | 0.04 | 3885.45 | 922 |
| e3-C | 77 | 98 | 10019 | 10805 | 0.08 | 10433 | 0.04 | **10401** | 0.04 | 3630.89 | 866 |
| e4-A | 77 | 98 | 6372 | 6489 | 0.02 | 6461 | 0.01 | 6547 | 0.02 | 4382.69 | 901 |
| e4-B | 77 | 98 | 8809 | 9216 | 0.05 | 9021 | 0.02 | 9214 | 0.05 | 1256.58 | 261 |
| e4-C | 77 | 98 | 11276 | 11824 | 0.05 | 11779 | 0.04 | 11883 | 0.05 | 4222.7 | 893 |
|  |  |  | Av.Dev(%): | | 3.84 | | 2.15 | | 2.94 | | |
|  |  |  | Nb hits: | | 0 | | 0 | | 0 | | |

## 4 Concluding Remarks and Future Research

This paper presents a resolution scheme for the CARP based on Ant Colony. The Ant Colony scheme is competitive with the best methods previously published providing high quality solutions in rather short computational time. It outperforms the CARPET algorithm and competes with the Genetic Algorithm for small and medium scale instances. The computational time is acceptable but the Ant Colony scheme can not compete, for a computational point of view, with the powerful Genetic Algorithm. This work is a step forward for the CARP resolution based on Ant Colony and proves Ant Colony Scheme competes with Taboo Search and Genetic Algorithms. It strengthens the previous published attempt of Doerner *et al.* [25]. To the best of our knowledge the proposed ACO is the first one proposed for the CARP providing high quality results for large scale instances. However, the performance of the proposed algorithm does not reach state-of-the-art performance and further researches are required to increase the convergence rate and to reduce the computational times.

# Index